\documentclass{article}

\usepackage{arxiv}

\usepackage[utf8]{inputenc} 
\usepackage[T1]{fontenc}    
\usepackage{hyperref}       
\usepackage{url}            
\usepackage{booktabs}       
\usepackage{amsfonts}       
\usepackage{nicefrac}       
\usepackage{microtype}      
\usepackage{amsmath}
\usepackage{lipsum}
\usepackage{graphicx}
\usepackage{subcaption}
\graphicspath{ {./images/} }
\usepackage{enumitem}
\setlist{itemsep=2pt, topsep=2pt}
\usepackage{xcolor}
\usepackage{algorithm}
\usepackage{algpseudocode}
\usepackage{multirow}
\usepackage{soul}

\DeclareMathOperator*{\argmin} {arg\,min}

\title{Neuroplasticity-inspired dynamic ANNs for multi-task demand forecasting}

\author{
 Mateusz Żarski \\
  Institute of Theoretical and Applied Informatics\\
  Polish Academy of Sciences\\
  Gliwice, Bałtycka 5, 44-100, Poland \\
  \texttt{mzarski@iitis.pl} \\
   \And
 Sławomir Nowaczyk \\
 School of Information Technology\\
  Halmstad University\\
  Kristian IV:s väg 3, 301 18 Halmstad, Sweden\\
  \texttt{slawomir.nowaczyk@hh.se} \\
}

\begin{document}
\maketitle
\begin{abstract}

This paper introduces a novel approach to Dynamic Artificial Neural Networks (D-ANNs) for multi-task demand forecasting called Neuroplastic Multi-Task Network (NMT-Net). Unlike conventional methods focusing on inference-time dynamics or computational efficiency, our proposed method enables structural adaptability of the computational graph during training, inspired by neuroplasticity as seen in biological systems. Each new task triggers a dynamic network adaptation, including similarity-based task identification and selective training of candidate ANN heads, which are then assessed and integrated into the model based on their performance. We evaluated our framework using three real-world multi-task demand forecasting datasets from Kaggle. We demonstrated its superior performance and consistency, achieving lower RMSE and standard deviation compared to traditional baselines and state-of-the-art multi-task learning methods.
NMT-Net offers a scalable, adaptable solution for multi-task and continual learning in time series prediction. The complete code for NMT-Net is available from our GitHub repository~\cite{Zarski_Multi_Forecasting_2025}. 

\end{abstract}


\section{Introduction}\label{sec:introduction}
Although Dynamic Artificial Neural Networks (D-ANNs) are not an entirely new concept in the field of Artificial Intelligence (AI)~\cite{shaw1997dynamic}, they have struggled to gain traction and become a cornerstone of the Deep Learning paradigm. Nevertheless, this promising branch of AI seems capable of redefining the field of neural algorithms in their scalability and multi-tasking capabilities~\cite{huang2024dynamic}. Somewhat surprisingly, recently defined dynamics of neural networks are usually limited to Sample-, Temporal-, or Spatial-wise Dynamics, where the networks' model itself is static, and dynamic behavior is provided with a dynamic inference graph~\cite{han2021dynamicneuralnetworkssurvey} or adaptive features and layers used~\cite{guo2024dynamic}. Moreover, currently the primary focus of dynamic neural network development lies in reducing computational costs~\cite{ICML} rather than improving Artificial Neural Networks (ANNs) performance in a multi-task setup.

This work proposes a novel approach to dynamic neural networks: Neuroplastic Multi-Task Network (NMT-Net). Our neural network model is dynamic with respect to the computational graph (and not the inference graph) in a multi-task, regression setup, resembling the phenomena of brain plasticity during task learning. With each new task, a neural network trained with our approach can update its computational graph by adding a new head to the graph to maximize its performance and simultaneously update the loss function according to this task.

We motivate our approach with a demand forecasting scenario.
Predicting product demand across multiple product categories and sales locations inherently results in a high-dimensional forecasting problem. This rapidly becomes challenging if all data are merged into a single predictive model, neglecting crucial heterogeneity. The result is model underperformance due to conflicting signals and variance across diverse demand profiles, as each product-location combination may exhibit unique temporal patterns, distribution characteristics, and demand dynamics influenced by local factors.
In contrast, naively treating each of these combinations as isolated tasks quickly becomes computationally infeasible while also ignoring valuable shared information, such as underlying demand drivers, market trends, or seasonal effects that manifest similarly across subsets of tasks.

A multi-task learning approach naturally addresses this tension by explicitly modeling shared structures through joint learning while still accommodating task-specific variations via dedicated parameters or adaptive mechanisms. Our work focuses on systematically leveraging shared latent representations to capture common predictive patterns while simultaneously preserving the ability to handle task-specific idiosyncrasies. However, finding the right balance between commonality and specificity is challenging, and D-ANNs appear to be the perfect tool for doing just that. We demonstrate robust performance improvements from NMT-Net in three regression problems using Kaggle public demand forecasting datasets.

The rest of the paper is organized as follows. Section~\ref{sec:related_works} describes existing research and approaches related to D-ANNs. Section~\ref{sec:method} provides a detailed description of our proposed NMT-Net method. 
Next, Sections~\ref{sec:exp_setup} and~\ref{sec:results} describe the experimental setup used to validate our approach and the results of the experiments that compare NMT-Net with baselines and state-of-the-art, respectively. 
Finally, in Section~\ref{sec:conclusions}, we present our final remarks and discuss possible future research directions.

\section{Related work}\label{sec:related_works}

There are three main topics present in the research literature that are relevant for the NMT-Net method: \textit{Dynamic Neural Networks}, \textit{Multi-task learning}, and the task of \textit{Demand forecasting}. 

\paragraph{Dynamic Neural Networks}\mbox{}\\

As mentioned in Section~\ref{sec:introduction}, the currently developed methods of implementing D-ANNs focus on Sample-, Temporal-, or Spatial-wise Network Dynamics. All these approaches remain similar due to the way the dynamics are implemented in the neural network model -- based on dynamic inference graphs~\cite{han2021dynamicneuralnetworkssurvey}. These methods are used primarily to reduce the computational cost of neural networks using techniques such as early exiting, layer skipping, and dynamic parameters~\cite{xu2022survey} \emph{etc.}, with the primary objective of reducing computational redundancy. They are used in tasks such as computer vision or text processing, where computational redundancy can play a significant role in the overall computational costs. 

Other approaches to D-ANNs include the use of Tapped-Delay Neural Network (TDNN) and Nonlinear autoregressive with exogenous inputs Network (NARX)~\cite{esposito2016dynamic,zika}. In this setup, the NN model implements network dynamics by introducing time series delays of the input or adding additional feedback connections, similar to the Recurrent Neural Network (RNN) architecture. This approach was successfully implemented in time series analysis~\cite{s18103408} and is closely related to the issue of demand forecasting.

Another approach to the problem of dynamic neural networks is to define computational and inference graphs during the network's training and implementation. DyNet~\cite{neubig2017dynet} authors proposed a framework for dynamic declaration and implementation of NN models, as opposed to a two-step declaration and execution present in other Deep Learning frameworks (\emph{e.g.}, TensorFlow). However, with this approach, the goal was again to reduce the computational complexity of the neural network. Other approaches to improve the performance of the ANN architecture involve using neural architecture search (NAS)~\cite{9508774}. These methods are used to automate the selection of a neural network architecture to best match the task it should perform -- \emph{e.g.}, in computer vision~\cite{SALEHIN202452}. In this field, frameworks have also been developed to apply NAS algorithms effectively~\cite{nas}.

\paragraph{Multi-task learning}\mbox{}\\

Multi-task learning aims to improve the performance and generalization of models and algorithms by simultaneously performing multiple tasks. Since its conception, it has been used for time-series prediction, multi-representation classification, sequential transfer~\cite{caruana1996algorithms,caruna2}, \emph{etc}. In previous applications of multi-task learning, various methods of designing or training ANNs were used, including hierarchical and tree-structured models~\cite{zhang2018overview}, task routing~\cite{crawshaw2020multi}, and feature sharing~\cite{ruder2017overview}, as well as other approaches like task relation or task decomposition~\cite{zhang2021survey}. 

Other multi-task learning methods that emerged recently include selective ANN branch sharing~\cite{pmlr-v119-guo20e}, knowledge distillation~\cite{distillation}, and adaptive model merging without relying on the initial training data~\cite{yang2024adamergingadaptivemodelmerging}. One of the most explored algorithm groups in multi-task learning is the grouping and clustering of tasks in the training dataset. Within this set of algorithms, methods for efficient task identification~\cite{NEURIPS2021_e77910eb,pmlr-v119-standley20a}, learning the task overlap~\cite{kumar2012learningtaskgroupingoverlap}, and task clustering~\cite{NIPS2008_fccb3cdc,zarski} can be distinguished. Closely related to this group are also research papers considering dynamic task prioritization~\cite{Guo_2018_ECCV} and multi-objective optimization~\cite{NEURIPS2018_432aca3a}.

Similarly to D-ANNs, there are also specific programming frameworks for multi-task problems, like MALSAR~\cite{zhou2011malsar}. It is developed for the MATLAB programming language and provides the user with a set of algorithms and regularization methods for multi-task problems. 


\paragraph{Demand forecasting}\mbox{}\\

Demand forecasting is a practical task that focuses on production, distribution, and ordering strategies as well as planning. Traditional methods such as Autoregressive Integrated Moving Average (ARIMA)~\cite{arima} and shallow learning models like Support Vector Machines~\cite{svm} have been widely used for time series forecasting and remain in practical use to this day. However, these techniques often struggle to capture complex, nonlinear patterns in data, especially in multi-task setups. 

Newer developments focus on the use of deep learning by leveraging ANN architectures that can model temporal dynamics within the time-series data, such as Long Short-Term Memory (LSTM)~\cite{LSTM} and Gated Recurrent Unit (GRU)~\cite{tian2024magru} models. Their popularity in this task is due to their ability to retain information over long sequences and manage temporal dependencies~\cite{jadli2022novel}, with the use of transformer models increasing for the same reason~\cite{zhang2024intermittent}.

Several studies have also explored hybrid approaches that combine ANNs with other machine learning techniques or numerical methods. For instance, models integrating Multi-Layer Perceptrons (MLPs) with autoregressive components have been investigated~\cite{Choi2009}, alongside hybrid models such as MLP-Fourier~\cite{fourier} and MLP-SVM~\cite{svm_hybrid}. Another notable direction in this research area involves incorporating trainable categorical embeddings within MLP architectures. These are particularly effective in scenarios involving highly differentiated products or when enhanced product categorization is required~\cite{Dholakia2020}.


\paragraph{Summary and research gap}\mbox{}\\

The topic of Dynamic Artificial Neural Networks has been explored from various perspectives, with most research primarily focusing on reducing computational complexity and improving the efficiency of time-series data processing. However, to the best of our knowledge, there are currently no effective methods for training ANNs with a dynamic computational graph. In the current work, we address this gap by proposing a novel approach tailored to a multi-task learning setup for demand forecasting. This enables dynamic adaptation of the ANN model's computational graph and dynamic task grouping as additional tasks are received during operation. With NMT-Net, we aim to expand the field of truly dynamic ANN models, inspired by brain plasticity phenomena in human and animal learning.

\section{Method}\label{sec:method}

This work proposes NMT-Net, a novel approach to Dynamic Artificial Neural Networks training in a multi-task setup, inspired by the mechanisms of neuroplasticity observed in the learning process of biological organisms~\cite{sagi2012learning,chen2022neuroplasticity}. Similarly to biological mechanisms, our approach assumes the following way a neural network learns new skills (\textit{tasks})~\cite{wenger2017expansion,kawai2015motor}:

\begin{enumerate}
    \item General neural structures are used for initial training on an unknown task
    \item New, specialized neural structures are temporarily created during learning to achieve high performance in the given task
    \item Acquired knowledge is transferred to another neural structure, either one responsible for similar previously learned tasks, or a unique structure fully responsible for the new task
    \item The temporary neural structure used during task-specific learning is released and available for re-use in the future
\end{enumerate}

To replicate this biological phenomenon as closely as practically possible, our NMT-Net method uses three key mechanisms during the training of consecutive tasks:

\begin{itemize}
    \item Similar task identification
    \item Temporary head training
    \item Head performance assessment and selection
\end{itemize}

\paragraph{Preliminaries and notations}\mbox{}\\

Datasets used in this study contain multiple regression tasks in the form of time series regarding the issue of demand forecasting. For each task, the dataset comprises a sequence of timesteps consisting of univariate vectors of categorical (product and vendor IDs) and continuous variables (\emph{e.g.}, the demand in a given period). We divided our dataset into three subsets, while preserving the order of the datapoints: 

\begin{itemize}
    \item $X_{pre}$ -- for initial pre-training phase of the model (obtaining initial model weights $\Theta_0$)
    \item $X_{post}$ -- for multi-learning phase 
    \item $X_{eval}$ -- for model evaluation
\end{itemize}

We used the ratio of $0.4:0.4:0.2$. In cases where the pre-training phase was not applicable, we used combined data $X_{train} = X_{pre} \cup X_{post}$ for model training.

\paragraph{Similar task identification}\mbox{}\\

In our setup, any given task consists of multiple training datapoints (sliding windows), each of a fixed $t$ number of time steps (lag). Task similarity identification is performed pairwise, between $n$ datapoints of feature vectors of a new, upcoming task $X_{new} = \{x_1^{(t)}, x_2^{(t)}, \ldots, x_n^{(t)}\}$ and a set of all tasks previously learned by the model $X_{old} = \{X_{1}, X_{2}, \ldots, X_{m}\}$. As the number of datapoints can be unequal across the tasks, for the similarity comparison, we use an average update to compute the average feature vector $\overline{X}^{(t)}$ for the whole set of datapoints for each task.

%

After obtaining average feature vector representations for the new task $X_{m+1}$, 
we use Root Mean Squared Error ($RMSE$) to compute a set of its pairwise task similarities against all average feature vector representations of the previously learned tasks $X_1, \ldots, X_m$. Thus, the already trained task $X_{sim}$ most closely related to the incoming task $X_{new}$ is defined as:

\begin{align}
X_{sim} = \argmin\Big( \{RMSE(\overline{X}_{new}, \overline{X}_{1}), RMSE(\overline{X}_{new}, \overline{X}_{2}), ..., RMSE(\overline{X}_{new}, \overline{X}_{m})\} \Big)
\label{eq:metric}
\end{align}

\paragraph{Temporary head training}\mbox{}\\

Given the new task $X_{new}$, and having selected the most similar, previously trained task $X_{sim}$, we then train two new temporary ANN heads -- one starting from an initial, general set of weights learned in the pre-training phase $\Theta_{0}$, 
and the second starting from the head trained using $X_{sim}$ data, $\Theta_{sim}$. 
The datasets used for the subsequent training of these two heads are as follows:
\begin{itemize}
    \item $\Theta_{0}$ is trained solely on $X_{new}$ data
    \item $\Theta_{sim}$ is trained on updated set of data: $X_{updated} = X_{sim} \cup X_{new}$
\end{itemize}
This phase aims to obtain the highest possible performing ANN head regarding the task at hand, $X_{new}$. However, if an existing head is to be used, its existing knowledge must be preserved to avoid catastrophic forgetting.

In other words, NMT-Net heads are designed to be re-used and perform multiple tasks as long as doing so is beneficial to the overall performance of the model. Thus, the number of tasks the ANN model knows is typically much higher than the number of ANN heads. The dynamic aspect of temporal head training allows the model to optimize its performance in the scope of a multi-task or continual learning setup.

\paragraph{Head performance assessment}\mbox{}\\

Given two temporary heads trained in the previous step, we assess their performance on the evaluation dataset for the new task $X_{new\_eval}$. With the regression objective, we use the training loss function $\mathcal{L}$ as an evaluation criterion~(\ref{eq:criterion}) for both heads and select the better-performing one as the new ANN head for the $X_{new}$ task:

\begin{equation}
    \Theta_{new} = \begin{cases} 
    \Theta_{0}, & \text{if } \mathcal{L}(\Theta_{0}(X_{new\_eval})) < \mathcal{L}(\Theta_{sim}(X_{new\_eval})) \\
    \Theta_{sim}, & \text{otherwise}.
    \end{cases}
    \label{eq:criterion}
\end{equation}

Subsequently, the better-performing head is saved, and the lower-scoring one is removed, thus releasing it from memory. Lastly, after adding the new task to the known task set, the model can be further updated with additional tasks in a multi-task or continual learning setup.

\begin{figure}[h]
    \centering
    \includegraphics[width=.92\linewidth]{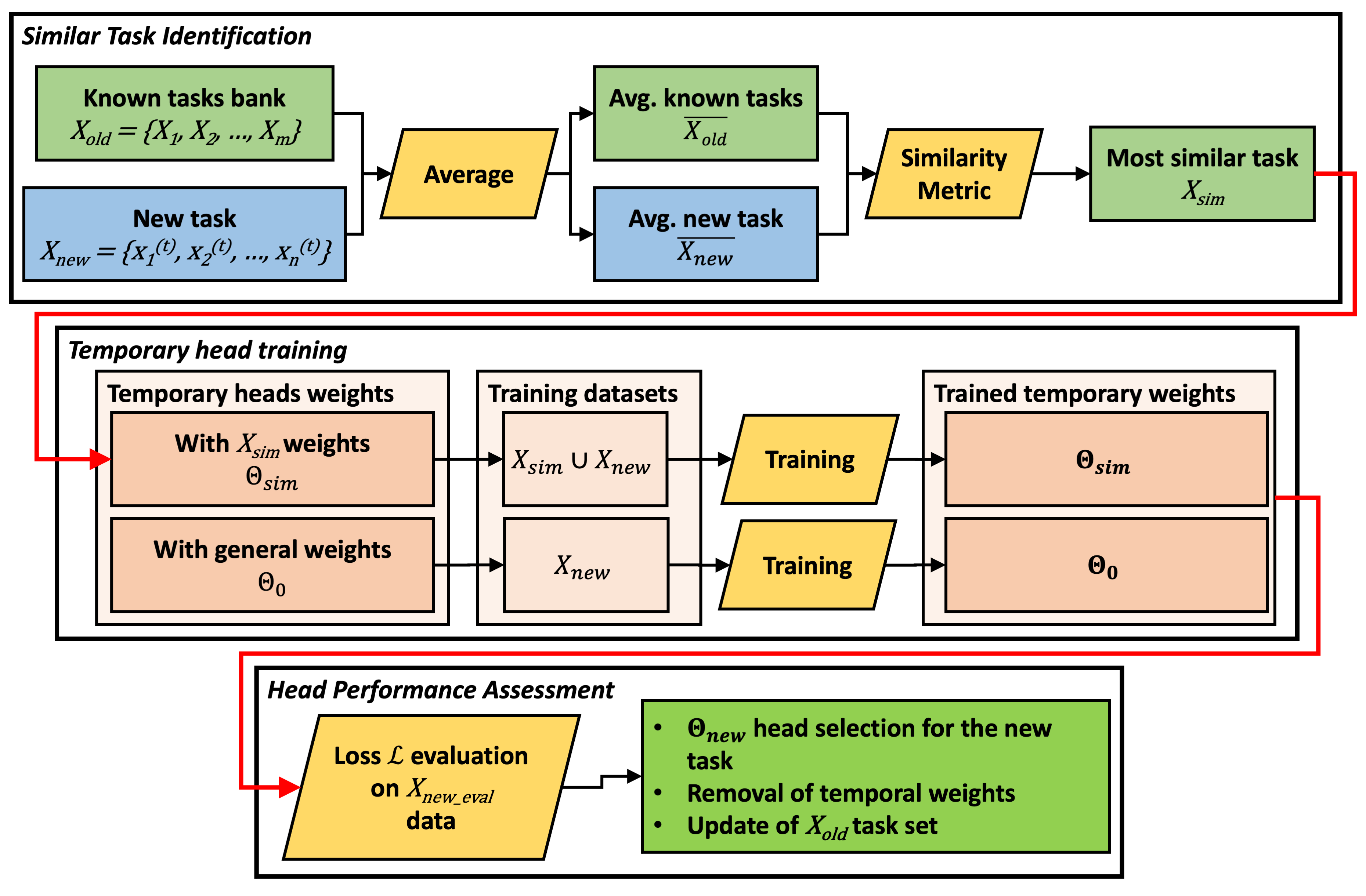}
    \caption{The process of dynamically adding a single task to the ANN model in a multi-task learning loop.}
    \label{fig:method}
\end{figure}

For clarity, the complete process of dynamic task addition to the NMT-Net model during its training cycle is depicted in Figure~\ref{fig:method} and formally defined in Algorithm~\ref{alg:main}.

Note that, unlike most contemporary approaches, NMT-Net does not use any hyperparameters to manage the training. This means that, apart from providing custom dataloaders and ANN architectures, our method is expected to work out-of-the-box for any further experiments. Furthermore, due to data-focused similarity search and extensive pre-training, our method can also be used for inference with previously unseen tasks; however, such experiments are out of scope for the current work.

\begin{algorithm}[]
  \caption{Main training loop}\label{alg:main}
  \begin{algorithmic}[1]
    \Require
      \Statex Model: $M$, with single $\Theta_0$ head
      \Statex Unknown task bank: $X_{unknown}$
      \Statex Similarity function: $sim$
      \Statex Loss function: $\mathcal{L}$

    \Ensure
      \State $X_{old} \gets \{\}$ \Comment{Known task bank is initially empty}

      \State $M.train(X_{unknown})$ \Comment{Pre-train model initially without task recognition}
      
      \While{$|X_{unknown}| \neq 0$}

      \State $X_{new} \gets random(X_{unknown})$ \Comment{Select new task for training}

      \State $X_{unknown} \gets X_{unknown} \setminus X_{new}$ \Comment{Update unknown task bank}

      \If{$|X_{old}| = 0$}

      \State $M.head_{1}.train(X_{new})$ \Comment{Train new ANN head on $X_{new}$ if it's the first head}
      \State $X_{old} \gets X_{old} \cup X_{new}$
      \State $\textbf{continue}$
      \EndIf

      \State $similarities \gets \{\}$
      \For{$X_{n}$ in $X_{old}$} \Comment{Compute similarities across known tasks}
        \State $similarities \gets similarities \cup sim(X_n, X_{new})$
      \EndFor

      \State $X_{sim} \gets \argmin(similarities)$ \Comment{Select index of the most similar task head for fine tuning}

      \State $M.\Theta_{sim}.train(X_{old}[X_{sim}] \cup X_{new})$ \Comment{Train new temporary heads}
      \State $M.\Theta_{0}.train(X_{new})$ 

      \If{$\mathcal{L}(M.\Theta_{sim}) < \mathcal{L}(M.\Theta_{0})$} \Comment{Assess temporary heads performance}

      \State $X_{old}[X_{sim}] \gets X_{old}[X_{sim}] \cup X_{new}$ \Comment{Append task to known task head}
      
      \Else

      \State $X_{old} \gets X_{old} \cup X_{new}$ \Comment{Add new task as head }

      \EndIf
      
      \EndWhile

  \end{algorithmic}
\end{algorithm}

\section{Experimental setup}\label{sec:exp_setup}

\subsection{Datasets used}

We used three datasets related to the demand forecasting task for our experimental setup, which are publicly available at Kaggle. We used \textit{Demand Forecasting (\textbf{DF})}~\cite{ds_1} dataset containing weekly sales from a three-year period, with multiple vendors and products, consisting of a total of 1155 tasks. The second dataset was a representative $10\%$ of the vast \textit{Store Item Demand Forecasting Challenge (\textbf{SIDF})}~\cite{ds_2} dataset, containing a total of 500 valid tasks based on vendor and product ID. Our last dataset was the \textit{Product Demand Forecast (\textbf{PDF})}~\cite{ds_3} dataset, which we divided into 93 tasks with product code, category, and warehouse. As an error function, for all our experiments, we used \textbf{RMSE} -- note that  RMSE (unlike \emph{e.g.} MSE) provides error in the same scale as the input data, so our results from Section~\ref{sec:results} cannot be compared head-to-head across the datasets. 

From the datasets tested, only the \textbf{DF} dataset contained additional information apart from the date, demand label, and categorical variables for product and vendor. Due to this, in order to remain consistent throughout our experiments, only categorical variables and demand labels were used for training. We also used a fixed number of lags for every experiment where applicable (with lag value $15$). We ensured that with every datapoint in training, the following data vector is preserved (in pre-training as well as single task training), totaling $17$ elements in the input vector:
\begin{align}
\texttt{input\_vector} &= \{vendor\_id, product\_id, damand_1, demand_2, ..., demand_{15}\} \label{eq:date_vector}
\end{align}


\subsection{Implementation}\label{ssec:implementation}

In our implementation, we used a Linux-based operating system and Python 3.11 environment with an Intel Xeon W-2255 CPU, 128 GB RAM, and an Nvidia RTX 6000 with 24 GB VRAM. The libraries we used in our code included \textit{scikit-Learn} and \textit{statsmodels} for our shallow ML experiments, as well as \textit{PyTorch} and \textit{PyTorch-Lightning} for ANN experiments.
Implementation of Algorithm~\ref{alg:main} is available in our repository~\cite{Zarski_Multi_Forecasting_2025}.

\subsection{Baselines}\label{ssec:baseline}

Our baseline experiments used ARIMA, Decision Trees, and Random Forests. The models were selected due to their consistent and extensive use in practical demand forecasting. 

Experiments on the Autoregressive Integrated Moving Average model were performed separately for each task to achieve the best performance. The data was considered as non-seasonal, with parameter grid search for all components of the model (AR (\textit{p}), MA (\textit{q}), and I (\textit{d})) resulting in a total of $135$ models per task. Below are the ranges used for every parameter:

\begin{itemize}
    \item $p = \{1, \dots, 15\}$, $step = 1$
    \item $q = \{1, \dots, 3\}$, $step = 1$
    \item $d = \{1, \dots, 3\}$, $step = 1$
\end{itemize}

Decision Trees and Random Forests experiments were also performed per task for consistency, with a set of grid-searched parameters for each algorithm run. The ranges of the parameters for DTs and RFs are listed below:

\begin{itemize}
    \item maximum regressor depth: $depth_{max} = \{4, \dots, 12\}$, $step=2$
    \item maximum number of leaf nodes: $leaf_{max} = \{10, 15, 20\}$
    \item maximum number of estimators (RFs only): $est_{max} = \{100, 150, 200, 250\}$
\end{itemize}

Additional experiments regarding post-pruning of the trained Decision Trees were also performed; however, the RMSE metric tended to be much worse, and ultimately, these experiments were excluded from the summary. 

\subsection{Multi-task learning SotA}\label{ssec:multi_tasking}

For our experiments covering the use of multi-tasking methods, we found four methods in recent literature against which we can compare NMT-Net. Two of them are multi-task regression models: Multi-Level Lasso (MTL)~\cite{lasso} and multi-task Wasserstein regularized regression (MTWR)~\cite{pmlr-v89-janati19a}, as shallow ML methods, as well as task grouping (TAG)~\cite{NEURIPS2021_e77910eb} and task clustering (MTL-cluster)~\cite{zarski}, as deep learning methods.

For our shallow ML experiments, we followed training recipes from the papers, with additional grid search over hyperparameters. For alpha parameters, we used a set $alphas=\{0.01, 0.05, 0.1, 0.2, 0.5, 1\}$, and for the number of relevant joint features for task training, we used $n_{features}=\{3,\dots,15\}$ with $step=1$.

For DL experiments, since the direct evaluation of different ANN model architectures suitable for the demand prediction task is outside the paper's scope, we used a simple MLP model with Cat2Vec categorical embedding for store and product IDs, along with a sequential model for past demands with a given lag. However, other ANN models, such as RNN or Transformer model architecture, can be used as a drop-in replacement for our NMT-Net method.

The ANN architecture consisted of two embedding layers performing categorical variables to 5-element vectors encoding, followed by concatenation with the rest of the input vector. Next, the sequential part of the network consisted of three MLP blocks: $\{Linear, ReLu, BatchNorm, Dropout\}$ with hidden dimensions $linear_{dim} = \{128, 256, 64\}$ and constant dropout rate of $0.5$. The head of the network was a single linear regression layer with a single neuron output for one-step forecasting. 

For optimization, the $AdamW$ optimizer was used with an initial learning rate of $0.01$ for the first $100$ pre-training epochs and $0.001$ for $50$ fine-tuning epochs. Learning rate was also scheduled with \textit{Reduce on Plateau} scheduler and the rate of $0.8$ for pre-training and $0.6$ for fine-tuning, as well as $20$ and $10$ step patience, respectively. RMSE loss was used as the loss function, and the model was trained with a consistent batch size of $5$ for every experiment performed.

\subsection{Neuroplastic Multi-Task Network (NMT-Net)}\label{ssec:dynamic_nn}

In order to stay consistent with our multi-tasking experiments, the overall model architecture remained the same for D-ANN experiments, with the only differences resulting from the dynamic nature of the training and the dynamic, multi-head capabilities of the model. For multi-head capabilities, we changed the last regression layer to \textit{layer dictionary} holding a set of regression layers along with their task-specific descriptions. This set is supplemented with additional regression layers during training, and pre-existing layers within it can also be fine-tuned during the training process with simultaneous optimizer update. The model also stores information on head-to-task relations.

For consistency with other MLP experiments, we also left the training process unchanged. Our optimizer, loss function, learning rates, and scheduler are the same as described in Subsection~\ref{ssec:multi_tasking}. 

\section{Results}\label{sec:results}
    
This section describes the experiments we performed with our chosen methods. The sections below provide a detailed description of each group of experiments, with methods, training recipes, and training parameters used. A summary of all performed experiments is included in Tab.~\ref{tab:results}. Note that the experiments were performed using a random seed and with at least five runs for every method. In the summary, we also include the standard deviation value ($\sigma$) for the results obtained.

In this section, in Tab.~\ref{tab:results}, we present the results of all the methods we tested. They are divided into three test datasets, and the primary measure we used is RMSE. We have also included in the table the standard deviation ($\sigma$) of the results obtained in the five randomized seed experiments. Later, in Subsection~\ref{ssec:additional}, we also include an ablation study, different similarity measurements, and additional information on head performance and task-to-head distribution.

\begin{table}[]
\caption{Results of experiments conducted on three test datasets}\label{tab:results}
\centering
\begin{tabular}{c|c|ccc}
\multirow{2}{*}{\textbf{Dataset}} & \multirow{2}{*}{\textbf{Method}}                          & \multicolumn{3}{c}{\textbf{RMSE}}            \\ \cline{3-5} 
          &                                                         & \multicolumn{1}{c}{\textbf{mean ($\sigma$)}} & \multicolumn{1}{c}{\textbf{min ($\sigma$)}} & \textbf{max ($\sigma$)} \\ \hline \hline
\textbf{} & ARIMA                                                   & 25.34 (4.35)                           & 3.43 (3.15)                           & 196.90 (7.22)    \\
          & DT                                                      & 28.98 (6.43)                           & 4.09 (3.11)                           & 360.56 (9.95)    \\
          & RF                                                      & 26.93 (5.55)                           & 3.32 (1.17)                           & 308.62 (9.21)    \\ \cline{2-5} 
          & MTL                                                      & 61.68 (4.03)                           & 56.65 (3.87)                           & 67.04 (5.17)    \\
\textbf{DF}          & MTWR                                                      & 45.58 (7.07)                             & 39.01 (4.22)                           & 56.65 (8.28)    \\ \cline{2-5} 
          & MTL-cluster & 18.48 (3.81)  & 8.72 (2.11)   & 69.19 (7.92) \\
          & TAG & 19.12 (3.33)                           & 8.09 (2.09)                           & 53.35 (6.39)     \\ \cline{2-5} 
          & \begin{tabular}[c]{@{}c@{}}\textbf{D-ANN}\\ \textbf{(ours)}\end{tabular} & \textbf{16.10 (0.21)}                                      & \textbf{3.25 (0.40)}                                     & \textbf{30.33 (1.01)}                \\ \hline \hline
\textbf{} & ARIMA                                                   & 12.43 (4.45)                           & 4.82 (6.36)                           & 23.61 (5.22)     \\
          & DT                                                      & 17.42 (7.27)                           & 5.45 (7.67)                           & 35.29 (7.78)     \\
          & RF                                                      & 17.40 (7.26)                           & 5.45 (7.22)                           & 35.30 (7.67)     \\ \cline{2-5} 
          & MTL                                                      & 33.45 (1.94)                           & 30.40 (1.21)                           & 35.62 (2.04)    \\
\textbf{SIDF}          & MTWR                                                      & 12.17 (5.26)                             & 8.60 (4.01)                           & 23.20 (5.56)    \\ \cline{2-5} 
          & MTL-cluster & 11.51 (2.92)  & 6.48 (2.17)   & 15.50 (2.66) \\
          & TAG & \textbf{6.64} (1.28)                           & 3.33 (1.36)                           & 14.24 (1.42)     \\ \cline{2-5} 
          & \begin{tabular}[c]{@{}c@{}}\textbf{D-ANN}\\ \textbf{(ours)}\end{tabular} & 6.66 \textbf{(0.17)}                                      & \textbf{3.22 (0.21)}                                     & \textbf{10.09 (0.53)}                \\ \hline \hline
\textbf{} & ARIMA                                                   & 6232 (1640)                            & 3.36 (142.0)                          & 8217 (1822)      \\
          & DT                                                      & 7294 (1945)                            & \textbf{2.71} (157.0)                          & 10095 (1874)     \\
          & RF                                                      & 7288 (1945)                            & \textbf{2.71} (157.0)                          & 10095 (1874)     \\ \cline{2-5} 
          & MTL                                          & 3833 (609.3)                           & 3715 (571.4)  & 3888 (621.4)    \\
\textbf{PDF}          & MTWR                                                      & 3383 (294.1)                             & 3070 (266.6)                           & 3942 (301.5)    \\ \cline{2-5} 
           & MTL-cluster & 1653 (431.9)  & 7.78 (3.34)   & 1992 (24.27) \\
                                  & TAG  & 858.2 (78.12) & 26.76 (45.47) & 2253 (167.2) \\ \cline{2-5} 
          & \begin{tabular}[c]{@{}c@{}}\textbf{D-ANN}\\ \textbf{(ours)}\end{tabular} & \textbf{573.5 (11.97)}                                      & 4.52 \textbf{(0.01)}                                     & \textbf{1893 (11.07)}               
\end{tabular}
\end{table}

As can be seen in the results, for the mean RMSE, our method outperformed other approaches in \textbf{DF} and \textbf{SIDF} datasets, while being close to the best-performing task grouping approach in \textbf{SIDF} dataset. Similarly, for minimum RMSE, our method again performed the best in two out of three datasets, with decision trees and random forests scoring better on \textbf{PDF} dataset. For maximum RMSE scores, the Dynamic ANNs outperformed all other methods, significantly lowering the score across all the datasets. Our method excelled the most in terms of the standard deviation of the experiments, having it lower mostly by an order of magnitude compared to other approaches, meaning that it remained the most consistent method across all the experiments. 

\begin{figure}[h]
    \centering
    \includegraphics[width=0.98\linewidth]{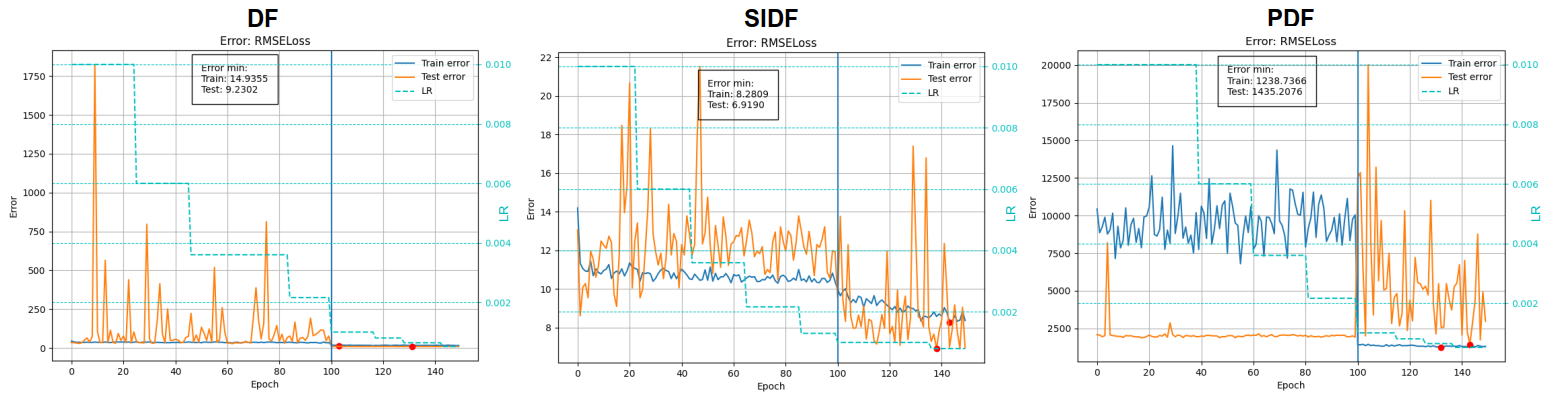}
    \caption{Sample training curves for the datasets tested.}
    \label{fig:curves}
\end{figure}

In Fig.~\ref{fig:curves}, we also included three sample training curves for the datasets used. In the figure, a vertical blue line separates the 100-epoch pre-training phase from the 50-epoch fine-tuning, head-specific phase. The main, black vertical axis contains RMSE measurements during training, and the dotted cyan axis contains the current learning rate for the epoch, also indicating when \textit{Reduce on Plateau} scheduler updated the learning rate. Also in the figure, two red dots indicate the lowest values for both training and testing errors, values of which are also shown in the \textit{Error min} box in the graphs. It can be seen that, depending on the dataset, the model training behaved differently, with \textbf{DF} model acting the most predictably, where training error remained lower than testing error throughout the training process, and \textbf{PDF} model learning the most head- (task group) specific behavior in the fine-tuning phase. It can also be seen that for the \textbf{SIDF} model, the fine-tuning phase helped significantly in the head-specific training phase, lowering both training and testing errors. Given the learning curves collected from all the experiments and datasets, it can be speculated that additional management of learning rate, batch size, and the length of each training phase could further improve our results.

\subsection{Additional experiments}\label{ssec:additional}

In this subsection, we cover additional experiments performed on the $X_{eval}$ part of the \textbf{DF} dataset. For all the additional experiments performed, we remained consistent with the training recipe described in Section~\ref{ssec:multi_tasking}, and at least five training runs were done for each experiment.

We performed an ablation study to assess the method of selecting the most similar, previously learned task to the task at hand, and also verified if the similarity calculation increases the model's performance at all. For the study, we selected four cases with varying similarity measurements:

\begin{itemize}
    \item Random (\textbf{RAND}) -- the task for fine-tuning is selected randomly from the $X_{old}$ set
    \item Median Absolute Error (\textbf{MedAE})
    \item Mean Gamma Deviance (\textbf{MGD})
    \item Root Mean Squared Error (\textbf{RMSE}) -- as in the experiments performed in Section~\ref{sec:results}
\end{itemize}

In Tab.~\ref{tab:ablation} we present the results of the ablation study. As can be seen in the results, using RMSE as a similarity metric still resulted in obtaining the best mean and minimum RMSE score, while obtaining a slightly lower maximum score. However, using two other metrics may result in obtaining a lower standard deviation across the scores obtained. As could be expected, using randomly selected model heads for fine-tuning resulted in a deterioration of the RMSE score, indicating that the model will benefit from jointly training the most similar tasks. 

\begin{table}[h]
\caption{Ablation experiments results}\label{tab:ablation}
\centering
\begin{tabular}{c|ccc}
\multirow{2}{*}{\textbf{\begin{tabular}[c]{@{}c@{}}SIM\\ Method\end{tabular}}} & \multicolumn{3}{c}{\textbf{RMSE}}                                            \\ \cline{2-4} 
                                                                               & \textbf{mean ($\sigma$)} & \textbf{min ($\sigma$)} & \textbf{max ($\sigma$)} \\ \hline \hline
RAND  & 18.04 (0.43) & 3.46 (0.23) & 29.76 (0.61) \\
MedAE & 16.40 (0.31) & 4.52 (\textbf{0.13}) & 31.03 (1.42) \\
MGD   & 16.24 \textbf{(0.16}) & 3.80 (0.47) & \textbf{29.11} (\textbf{0.57}) \\
RMSE  & \textbf{16.10} (0.21) & \textbf{3.25} (0.40) & 30.33 (1.01)
\end{tabular}
\end{table}

For our next experiments, we dive deeper into the process of training the models to assess how well each SIM function behaves as the task number increases. In order to do so, we evaluated the performance and characteristics of heads added during the training for each of the similarity metrics. We specified three additional characteristics that we measured during the course of multi-task training:

\begin{itemize}
    \item Total head count
    \item The number of tasks performed by a single head 
    \item The performance of heads across the added tasks
\end{itemize}

\begin{figure}[htbp]
    \centering

    \begin{subfigure}{\linewidth}
        \centering
        \includegraphics[width=0.95\linewidth]{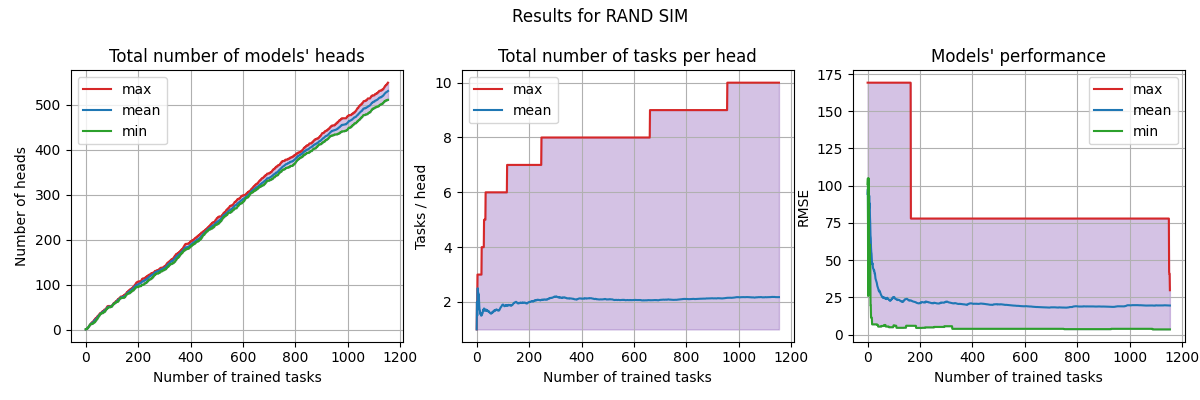}
        \label{fig:fig1}
    \end{subfigure}


    \begin{subfigure}{\linewidth}
        \centering
        \includegraphics[width=0.95\linewidth]{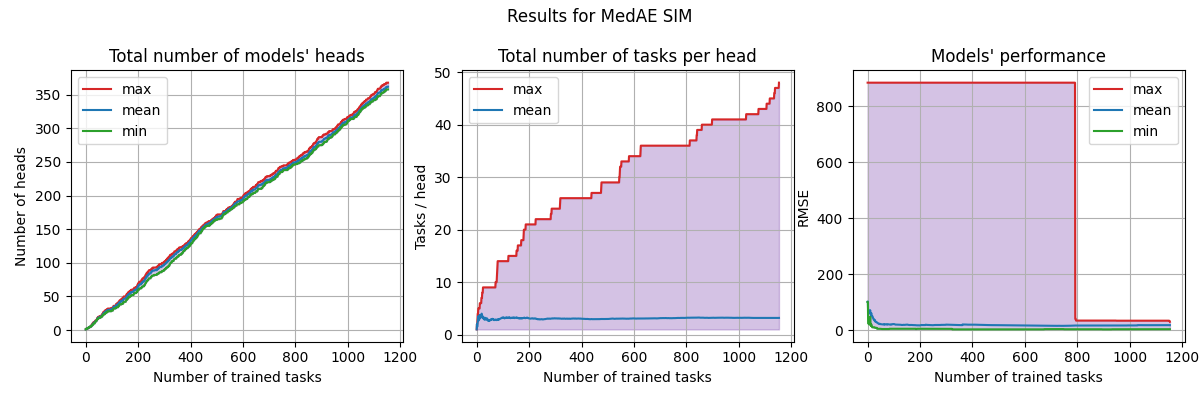}
        \label{fig:fig2}
    \end{subfigure}


    \begin{subfigure}{\linewidth}
        \centering
        \includegraphics[width=0.95\linewidth]{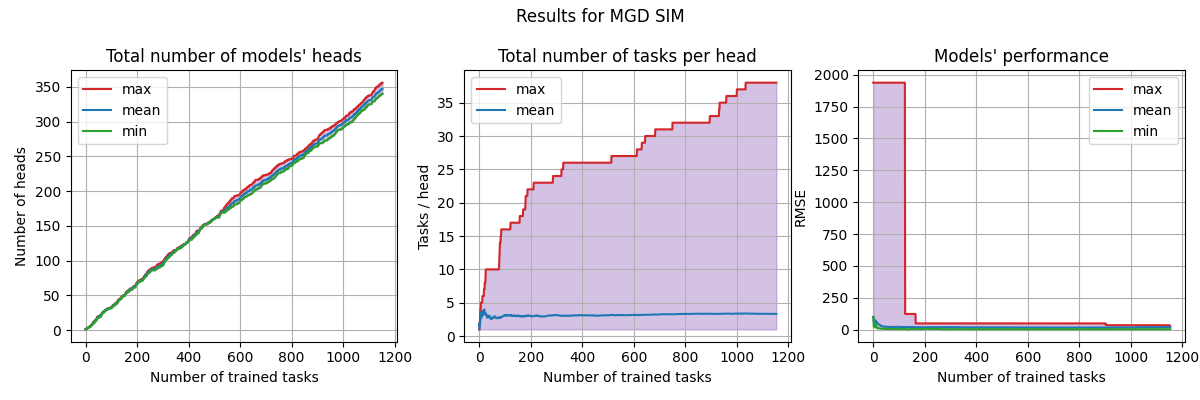}
        \label{fig:fig3}
    \end{subfigure}


    \begin{subfigure}{\linewidth}
        \centering
        \includegraphics[width=0.95\linewidth]{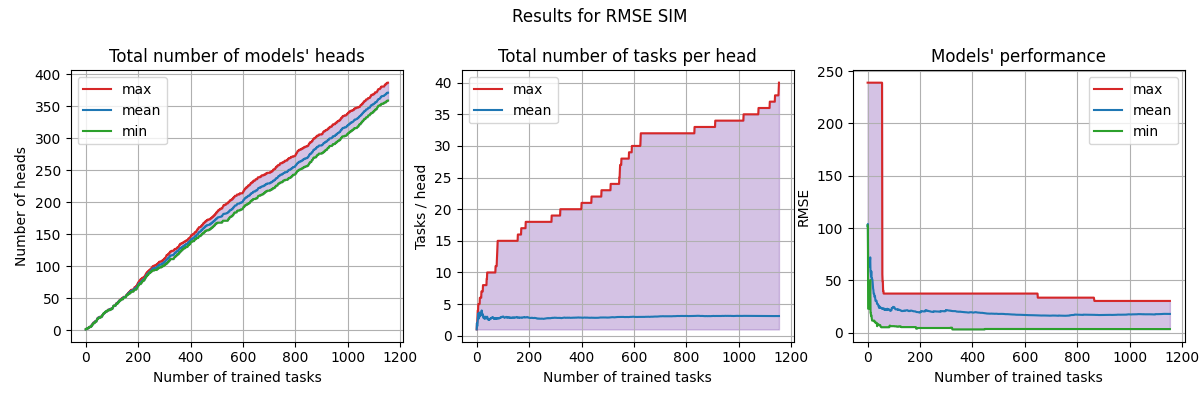}
        \label{fig:fig4}
    \end{subfigure}

    \caption{The results for the total number of models' heads, the number of tasks performed per head, and overall models' performance for the different SIM methods.}
    \label{fig:add}
\end{figure}

In Fig.~\ref{fig:add}, we present the additional models' characteristics for every SIM method we tested. For the first column -- the total number of models' heads, it can be seen that all three SIM methods (MedAE, MGD, and RMSE) maintain similar head count for the given task number, while for the RAND method total head count is significantly higher, indicating that the tasks could not be grouped in a effective manner. 

The same can be seen in the next column -- the total number of tasks per head. In this column, we present the maximum and mean number of tasks per head in the network for all experiments performed. We omitted the minimum value, as the number is consistently $1$ every time a new head is added. Once again, apart from the RAND method, the MedAE, MGD, and RMSE methods stay close to each other, with the average tasks per head being approximately $4$. For the RAND method, a much lower task count per head can be observed, indicating that the tasks that were trained together did not significantly improve the performance of the model. 

Lastly, in the third column, we show the overall model performance graphs as they changed during the addition of the tasks. The most significant difference can be seen in the maximum RMSE score across the SIM methods. While for the RAND method, the maximum RMSE started with the lowest value, the model then struggled to improve it for the whole course of training, meaning that the addition of new tasks did little to improve the overall performance. Similarly, the MedAE method also struggled to improve over time, and only with the addition of task $800$ could it improve the performance. The best-performing methods were the MGD and RMSE, where, thanks to the similarity found in the task, they improved the worst-performing heads with tasks $168$ and $73$, respectively. It can also be noted that both the MGD and RMSE methods improved the worst-performing heads the most frequently -- $3$ times, while the MedAE method could provide the improvement only once. 

Considering our findings from these additional experiments, it can be seen that using RMSE as the similarity measurement method yields the best results in multi-task scenarios. However, it is also worth noting that alternative methods to the ones presented (\emph{e.g.} similarity based on the trained models' weights rather than task similarity) can also be valid. We also conclude that:

\begin{enumerate}
    \item Uncovering the task similarity can significantly improve the performance of the multi-task model.
    \item When grouped, similar tasks will improve the overall performance of the model.
    \item Using the different SIM methods has a notable impact on how well the tasks are grouped in the scope of both model size and performance.
    \item The rate of model performance improvements is two-way dependent on similarity measurements, as it matters for both task distribution across models' heads and the similarity of the upcoming task to the tasks already known by the model.
\end{enumerate}

\section{Conclusions and future work}\label{sec:conclusions}

In this work, we proposed a novel method for Dynamic Artificial Neural Networks aimed at enhancing multi-task and continual learning models by introducing structural adaptability to the neural architecture during training. Our method, inspired by neuroplasticity as seen in the task learning of biological organisms, allows for dynamically adapting the computational graph through task similarity assessment, temporary ANN head training, and selective head integration. Our approach proved to allow structural flexibility, significantly improving performance and consistency across various multi-task problems. We firmly believe that our method can also be applied to other multi-task domains, as well as to continual learning problems, including reinforcement learning setups.

Experimental results on three public datasets demonstrated that our method outperforms traditional models such as ARIMA, Decision Trees, and Random Forests, as well as modern multi-task learning strategies (task grouping and task clustering). In particular, our approach achieved the lowest RMSE and standard deviation across most experiments, indicating both accuracy and robustness.

With our promising results, several research paths are open for further investigation. As stated in Section~\ref{sec:results}, additional, dynamic hyperparameter tuning during the training process can further improve the performance of our method. We also see potential for integrating our method with other ANN architectures, such as RNNs, CNNs, and Transformer models, and testing our approach with different types of data and tasks. 

Other future research opportunities we consider valid are employing our method for continual learning in the scope of reinforcement learning setups, as our method was inspired by reinforcement learning-type tasks in biological environments. Using our method on Atari, MuJoCo, or MiniHack environments could result in improvements of the current state-of-the-art models, and also allow us to compare our findings on datasets that are more widely used for the multi-task setups than the task of demand forecasting.

Lastly, with the core of our method tested and verified, we want to delve into more complex methods of integrating into ANNs' structural models during training. In this paper, we presented task-group-specific computational model dynamics focusing on the model's head layers. In our further work, we want to tackle the problem of modifying specific ANN layers in the global scope in order to achieve fully cross-model dynamic computational ANN graphs.

Our method and accompanying codebase, available on GitHub, offer a solid foundation for further development of dynamic, neuroplasticity-inspired architectures in multi-task learning scenarios. We hope this work encourages future research into biologically plausible and structurally adaptive AI systems, as we will pursue it onward.

\footnotesize
\bibliographystyle{splncs04}  

\bibliography{bib}

\begin{thebibliography}{10}
\providecommand{\url}[1]{\texttt{#1}}
\providecommand{\urlprefix}{URL }
\providecommand{\doi}[1]{https://doi.org/#1}

\bibitem{ICML}
Dynamic neural networks. ICML Workshop  (2022), \url{https://icml.cc/virtual/2022/workshop/13451}, available at \url{https://icml.cc/virtual/2022/workshop/13451}

\bibitem{s18103408}
Adeyemi, O., Grove, I., Peets, S., Domun, Y., Norton, T.: Dynamic neural network modelling of soil moisture content for predictive irrigation scheduling. Sensors  \textbf{18}(10) (2018). \doi{10.3390/s18103408}, \url{https://www.mdpi.com/1424-8220/18/10/3408}

\bibitem{zika}
Akhtar, M., Kraemer, M.U.G., Gardner, L.M.: A dynamic neural network model for predicting risk of zika in real time. BMC Medicine  (2019). \doi{10.1186/s12916-019-1389-3}

\bibitem{caruana1996algorithms}
Caruana, R.: Algorithms and applications for multitask learning. In: ICML. pp. 87--95. Citeseer (1996)

\bibitem{caruna2}
Caruana, R.: Multitask learning. Machine Learning  (1997). \doi{10.1023/A:1007379606734}

\bibitem{LSTM}
Chandriah, K.K., Naraganahalli, R.: Rnn / lstm with modified adam optimizer in deep learning approach for automobile spare parts demand forecasting. Multimedia Tools and Applications  (2021). \doi{10.1007/s11042-021-10913-0}

\bibitem{chen2022neuroplasticity}
Chen, S.A., Goodwill, A.M.: Neuroplasticity and adult learning. In: Third international handbook of lifelong learning, pp. 1--19. Springer (2022)

\bibitem{Choi2009}
Choi, G.S., Yu, C., Jin, R.M., Yu, S.K., Chun, M.G.: Short-term water demand forecasting algorithm using ar model and mlp. Journal of Korean Institute of Intelligent Systems  \textbf{19}(5),  713–719 (2009). \doi{10.5391/jkiis.2009.19.5.713}

\bibitem{crawshaw2020multi}
Crawshaw, M.: Multi-task learning with deep neural networks: A survey. arXiv preprint arXiv:2009.09796  (2020)

\bibitem{Dholakia2020}
Dholakia, R., Randeria, R., Dholakia, R., Ashar, H., Rana, D.: Cognitive Demand Forecasting with Novel Features Using Word2Vec and Session of the Day, pp. 59--72. Springer International Publishing, Cham (2020). \doi{10.1007/978-3-030-38445-6\_5}

\bibitem{esposito2016dynamic}
Esposito, E., De~Vito, S., Salvato, M., Bright, V., Jones, R.L., Popoola, O.: Dynamic neural network architectures for on field stochastic calibration of indicative low cost air quality sensing systems. Sensors and Actuators B: Chemical  \textbf{231},  701--713 (2016)

\bibitem{arima}
Fattah, J., Ezzine, L., Aman, Z., Moussami, H.E., Lachhab, A.: Forecasting of demand using arima model. International Journal of Engineering Business Management  \textbf{10},  1847979018808673 (2018). \doi{10.1177/1847979018808673}, \url{https://doi.org/10.1177/1847979018808673}

\bibitem{NEURIPS2021_e77910eb}
Fifty, C., Amid, E., Zhao, Z., Yu, T., Anil, R., Finn, C.: Efficiently identifying task groupings for multi-task learning. In: Ranzato, M., Beygelzimer, A., Dauphin, Y., Liang, P., Vaughan, J.W. (eds.) Advances in Neural Information Processing Systems. vol.~34, pp. 27503--27516. Curran Associates, Inc. (2021), \url{https://proceedings.neurips.cc/paper_files/paper/2021/file/e77910ebb93b511588557806310f78f1-Paper.pdf}

\bibitem{guo2024dynamic}
Guo, J., Chen, C.P., Liu, Z., Yang, X.: Dynamic neural network structure: A review for its theories and applications. IEEE Transactions on Neural Networks and Learning Systems  (2024)

\bibitem{Guo_2018_ECCV}
Guo, M., Haque, A., Huang, D.A., Yeung, S., Fei-Fei, L.: Dynamic task prioritization for multitask learning. In: Proceedings of the European Conference on Computer Vision (ECCV) (September 2018)

\bibitem{pmlr-v119-guo20e}
Guo, P., Lee, C.Y., Ulbricht, D.: Learning to branch for multi-task learning. In: III, H.D., Singh, A. (eds.) Proceedings of the 37th International Conference on Machine Learning. Proceedings of Machine Learning Research, vol.~119, pp. 3854--3863. PMLR (13--18 Jul 2020), \url{https://proceedings.mlr.press/v119/guo20e.html}

\bibitem{han2021dynamicneuralnetworkssurvey}
Han, Y., Huang, G., Song, S., Yang, L., Wang, H., Wang, Y.: Dynamic neural networks: A survey (2021), \url{https://arxiv.org/abs/2102.04906}

\bibitem{huang2024dynamic}
Huang, G.: Dynamic neural networks: advantages and challenges. National Science Review  \textbf{11}(8),  nwae088 (2024)

\bibitem{NIPS2008_fccb3cdc}
Jacob, L., Vert, J.p., Bach, F.: Clustered multi-task learning: A convex formulation. In: Koller, D., Schuurmans, D., Bengio, Y., Bottou, L. (eds.) Advances in Neural Information Processing Systems. vol.~21. Curran Associates, Inc. (2008), \url{https://proceedings.neurips.cc/paper_files/paper/2008/file/fccb3cdc9acc14a6e70a12f74560c026-Paper.pdf}

\bibitem{jadli2022novel}
Jadli, A., Lahmer, E.H.B., et~al.: A novel lstm-gru-based hybrid approach for electrical products demand forecasting. International Journal of Intelligent Engineering \& Systems  \textbf{15}(3) (2022)

\bibitem{pmlr-v89-janati19a}
Janati, H., Cuturi, M., Gramfort, A.: Wasserstein regularization for sparse multi-task regression. In: Chaudhuri, K., Sugiyama, M. (eds.) Proceedings of the Twenty-Second International Conference on Artificial Intelligence and Statistics. Proceedings of Machine Learning Research, vol.~89, pp. 1407--1416. PMLR (16--18 Apr 2019), \url{https://proceedings.mlr.press/v89/janati19a.html}

\bibitem{nas}
Jin, H., Song, Q., Hu, X.: Auto-keras: An efficient neural architecture search system. In: Proceedings of the 25th ACM SIGKDD International Conference on Knowledge Discovery \& Data Mining. p. 1946–1956. KDD '19, Association for Computing Machinery, New York, NY, USA (2019). \doi{10.1145/3292500.3330648}, \url{https://doi.org/10.1145/3292500.3330648}

\bibitem{ds_3}
Kaggle: Product demand forecast (Aug 2017), \url{https://www.kaggle.com/code/flavioakplaka/product-demand-forecast}

\bibitem{ds_2}
Kaggle: Store item demand forecasting challenge (Jun 2018), \url{https://www.kaggle.com/competitions/demand-forecasting-kernels-only}

\bibitem{kawai2015motor}
Kawai, R., Markman, T., Poddar, R., Ko, R., Fantana, A.L., Dhawale, A.K., Kampff, A.R., {\"O}lveczky, B.P.: Motor cortex is required for learning but not for executing a motor skill. Neuron  \textbf{86}(3),  800--812 (2015)

\bibitem{svm_hybrid}
Kilimci, Z.H., Akyuz, A.O., Uysal, M., Akyokus, S., Uysal, M.O., Atak~Bulbul, B., Ekmis, M.A.: An improved demand forecasting model using deep learning approach and proposed decision integration strategy for supply chain. Complexity  \textbf{2019}(1),  9067367 (2019). \doi{10.1155/2019/9067367}

\bibitem{kumar2012learningtaskgroupingoverlap}
Kumar, A., III, H.D.: Learning task grouping and overlap in multi-task learning (2012), \url{https://arxiv.org/abs/1206.6417}

\bibitem{distillation}
Li, W.H., Bilen, H.: Knowledge distillation for multi-task learning. In: Bartoli, A., Fusiello, A. (eds.) Computer Vision -- ECCV 2020 Workshops. pp. 163--176. Springer International Publishing, Cham (2020)

\bibitem{9508774}
Liu, Y., Sun, Y., Xue, B., Zhang, M., Yen, G.G., Tan, K.C.: A survey on evolutionary neural architecture search. IEEE Transactions on Neural Networks and Learning Systems  \textbf{34}(2),  550--570 (2023). \doi{10.1109/TNNLS.2021.3100554}

\bibitem{lasso}
Lozano, A.C., Swirszcz, G.: Multi-level lasso for sparse multi-task regression. In: Proceedings of the 29th International Coference on International Conference on Machine Learning. p. 595–602. ICML'12, Omnipress, Madison, WI, USA (2012)

\bibitem{neubig2017dynet}
Neubig, G., Dyer, C., Goldberg, Y., Matthews, A., Ammar, W., Anastasopoulos, A., Ballesteros, M., Chiang, D., Clothiaux, D., Cohn, T., et~al.: Dynet: The dynamic neural network toolkit. arXiv preprint arXiv:1701.03980  (2017)

\bibitem{fourier}
Odan, F.K., Reis, L.F.R.: Hybrid water demand forecasting model associating artificial neural network with fourier series. Journal of Water Resources Planning and Management  \textbf{138}(3),  245--256 (2012). \doi{10.1061/(ASCE)WR.1943-5452.0000177}

\bibitem{ds_1}
Rao, A.V.S.: Demand forecasting dataset (Jul 2020), \url{https://www.kaggle.com/datasets/aswathrao/demand-forecasting}

\bibitem{ruder2017overview}
Ruder, S.: An overview of multi-task learning in deep neural networks. arXiv preprint arXiv:1706.05098  (2017)

\bibitem{sagi2012learning}
Sagi, Y., Tavor, I., Hofstetter, S., Tzur-Moryosef, S., Blumenfeld-Katzir, T., Assaf, Y.: Learning in the fast lane: new insights into neuroplasticity. Neuron  \textbf{73}(6),  1195--1203 (2012)

\bibitem{SALEHIN202452}
Salehin, I., Islam, M.S., Saha, P., Noman, S., Tuni, A., Hasan, M.M., Baten, M.A.: Automl: A systematic review on automated machine learning with neural architecture search. Journal of Information and Intelligence  \textbf{2}(1),  52--81 (2024). \doi{https://doi.org/10.1016/j.jiixd.2023.10.002}, \url{https://www.sciencedirect.com/science/article/pii/S2949715923000604}

\bibitem{NEURIPS2018_432aca3a}
Sener, O., Koltun, V.: Multi-task learning as multi-objective optimization. In: Bengio, S., Wallach, H., Larochelle, H., Grauman, K., Cesa-Bianchi, N., Garnett, R. (eds.) Advances in Neural Information Processing Systems. vol.~31. Curran Associates, Inc. (2018), \url{https://proceedings.neurips.cc/paper_files/paper/2018/file/432aca3a1e345e339f35a30c8f65edce-Paper.pdf}

\bibitem{shaw1997dynamic}
Shaw, A.M., Doyle~III, F.J., Schwaber, J.S.: A dynamic neural network approach to nonlinear process modeling. Computers \& chemical engineering  \textbf{21}(4),  371--385 (1997)

\bibitem{pmlr-v119-standley20a}
Standley, T., Zamir, A., Chen, D., Guibas, L., Malik, J., Savarese, S.: Which tasks should be learned together in multi-task learning? In: III, H.D., Singh, A. (eds.) Proceedings of the 37th International Conference on Machine Learning. Proceedings of Machine Learning Research, vol.~119, pp. 9120--9132. PMLR (13--18 Jul 2020), \url{https://proceedings.mlr.press/v119/standley20a.html}

\bibitem{tian2024magru}
Tian, R., Wang, B., Wang, C.: Magru: Multi-layer attention with gru for logistics warehousing demand prediction. KSII Transactions on Internet \& Information Systems  \textbf{18}(3) (2024)

\bibitem{wenger2017expansion}
Wenger, E., Brozzoli, C., Lindenberger, U., L{\"o}vd{\'e}n, M.: Expansion and renormalization of human brain structure during skill acquisition. Trends in cognitive sciences  \textbf{21}(12),  930--939 (2017)

\bibitem{xu2022survey}
Xu, C., McAuley, J.: A survey on dynamic neural networks for natural language processing. arXiv preprint arXiv:2202.07101  (2022)

\bibitem{yang2024adamergingadaptivemodelmerging}
Yang, E., Wang, Z., Shen, L., Liu, S., Guo, G., Wang, X., Tao, D.: Adamerging: Adaptive model merging for multi-task learning (2024), \url{https://arxiv.org/abs/2310.02575}

\bibitem{svm}
Yue, L., Yafeng, Y., Junjun, G., Chongli, T.: Demand forecasting by using support vector machine. In: Third International Conference on Natural Computation (ICNC 2007). vol.~3, pp. 272--276 (2007). \doi{10.1109/ICNC.2007.324}

\bibitem{zarski}
{\.{Z}}arski, M., Nowaczyk, S.: Balancing performance and scalability of demand forecasting ml models. In: Krempl, G., Puolam{\"a}ki, K., Miliou, I. (eds.) Advances in Intelligent Data Analysis XXIII. pp. 127--140. Springer Nature Switzerland, Cham (2025)

\bibitem{zhang2024intermittent}
Zhang, G.P., Xia, Y., Xie, M.: Intermittent demand forecasting with transformer neural networks. Annals of Operations Research  \textbf{339}(1),  1051--1072 (2024)

\bibitem{zhang2018overview}
Zhang, Y., Yang, Q.: An overview of multi-task learning. National Science Review  \textbf{5}(1),  30--43 (2018)

\bibitem{zhang2021survey}
Zhang, Y., Yang, Q.: A survey on multi-task learning. IEEE transactions on knowledge and data engineering  \textbf{34}(12),  5586--5609 (2021)

\bibitem{zhou2011malsar}
Zhou, J., Chen, J., Ye, J.: Malsar: Multi-task learning via structural regularization. Arizona State University  \textbf{21},  1--50 (2011)

\bibitem{Zarski_Multi_Forecasting_2025}
Żarski, M.: {Multi\_Forecasting} (Jun 2025), \url{https://github.com/MatZar01/Multi_Forecasting}

\end{thebibliography}

\end{document}